%% file: main.tex
\documentclass[sigconf]{acmart-me}
\usepackage[utf8]{inputenc}
\usepackage{graphicx}
\usepackage{subcaption}
\usepackage[acronym]{glossaries}
\usepackage{booktabs}
\usepackage{float}
\usepackage{graphicx}
\usepackage{balance}

\newacronym{dl}{DL}{Deep Learning}
\newacronym{gi}{GI}{Gastrointestinal Tract}
\newacronym{gt}{GT}{Ground Truth}
\newacronym{std}{std}{Standard Deviation}
\newacronym{miou}{mIoU}{Mean Intersection over Union}
\newacronym{dc}{DC}{Dice Coefficient}

\title{Pyramid-Focus-Augmentation: Medical Image Segmentation with Step-Wise Focus}
\author{Vajira Thambawita\textsuperscript{1,2}, Steven Hicks\textsuperscript{1,2},  P{\aa}l Halvorsen\textsuperscript{1,2}, Michael A. Riegler\textsuperscript{1} \\}
\affiliation{
\textsuperscript{1}SimulaMet, Norway \ \ \ {}
\textsuperscript{2}Oslo Metropolitan University, Norway\ \ \ \
}
\email{Contact: vajira@simula.no}
\date{October 2020}

\setcopyright{rightsretained}
\renewcommand{\shortauthors}{Thambawita et al.}

\acmDOI{}

\acmISBN{}

\acmConference[MediaEval'20]{Multimedia Evaluation Workshop}{December 14-15 2020}{Online} 
\acmYear{2020}
\copyrightyear{}

\acmPrice{}

\begin{document}

\begin{abstract}
Segmentation of findings in the gastrointestinal tract is a challenging but also important task which is an important building stone for sufficient automatic decision support systems. In this work, we present our solution for the Medico 2020 task, which focused on the problem of colon polyp segmentation. We present our simple but efficient idea of using an augmentation method that uses grids in a pyramid-like manner (large to small) for segmentation. Our results show that the proposed methods work as indented and can also lead to comparable results when competing with other methods.
\end{abstract}

\maketitle

\section{Introduction}

Segmented polyp regions in \gls{gi} images~\cite{8512197} can provide detailed analysis to doctors to identify correct areas to proceed with treatments compared to other computer-aided analysis such as classification~\cite{acm_vajira, thambawita2018medico, hicks2018deep} and detection~\cite{polyp_detection} which provide less detailed information about the exact region and size of the affected area. However, training \gls{dl} models to perform segmentation for medical data is challenging because of the lack of medical domain images as a result of tight privacy restrictions, the high cost for annotating medical data using experts, and a lower number of true positive findings compared to true negatives. In this paper, we present our approach for the participation in the 2020 Medico Segmentation Challenge~\cite{jha2020medico},  for which we introduce a novel augmentation technique called \textbf{pyramid-focus-augmentation (PYRA)}. PYRA can be used to improve the performance of segmentation tasks when we have a small dataset to train our \gls{dl} models or if the number of positive findings is small. Further, our method can focus doctors' attention to regions of polyps gradually. In addition to that the output of the method is also adjustable meaning, we could present a lower resolution of the grid if this is sufficient for the task at hand which can help to save processing time. Finally, our technique can also be applied to any segmentation task using any deep learning segmentation model.

\section{Method}
Our method has two main steps: data augmentation with PYRA using pre-defined grid sizes followed by training of a \gls{dl} model with the resulting augmented data. The source code for our method can be found in our GitHub\footnote{GitHub: \url{https://vlbthambawita.github.io/PYRA/}} repository.
The development dataset~\cite{jha2020kvasir} provided by the organizers has $ 1000 $ polyp images with corresponding ground truth masks.  We divided it into two parts such that $ 800 $ images are used for model training and $ 200 $ for testing.

\subsection{PYRA Data Augmentation}
As the first step in PYRA, we generate checker board grids as illustrated in the first row of Figure~\ref{fig:main_grid} with sizes of $ N\times N $ with $N$ values of $4,8,16,32,64,128$ and $256$. $ N $ should be selected such that $ image\_size\:\%\:N = 0 $. Applying these eight grid augmentations to the training dataset with $ 800 $ images increases the training data to $ 800 \times 8 = 6400 $ images. 

For the second step, we convert the \gls{gt} segmentation masks  into a grid-based representation of the \gls{gt} corresponding to the grid sizes. For example, if the grid size is $ 8 \times 8 $, then the corresponding \gls{gt} is a $ 8 \times 8$ converted \gls{gt}. 

The transformation of the ground truth masks to gridded masks is performed as following: (i) we divide the gt into the input grid size, (ii) we counted true pixels of each grid cell, (iii) if the number of true pixels is larger than $ 0 $, we converted the whole cell into a true cell. An example of a converted \gls{gt} is depicted on the top of Figure \ref{fig:model}.

\subsection{Experimental Setup and Model training}
\input{images/model}
\input{images/main_grid}

We have set up four experiments: Exp-1, Exp-2, Exp-3, and Exp-4 to show the performance of PYRA. Exp-1 and Exp-2 represent two baseline experiments. Exp-1 uses only the $ 800 $ training images without any augmentations. In Exp-2, we used general augmentations such as Affine, Coarse Dropout, and Additive Gaussian Noise from the library called \textit{imgaug}~\cite{imgaug}. Exp-3 and Exp-4 are using our PYRA with the data from Exp-1 and Exp-2, respectively. The training dataset size was changed from $ 800 $ to $ 6400 $ after applying PYRA. However, we validated our experiments only using $ 200 $ images reserved for testing. We have used one data loader for all experiments to maintain a fair evaluation. The baseline experiments Exp-1 and Exp-2 used the data loader with a grid size of $256 \times 256 $ which represents the original \gls{gt} masks without any conversion.

We have used the Unet architecture~\cite{ronneberger2015u} as our \gls{dl} model to perform the polyp segmentation task. We trained the Unet model with a stacked input using a polyp image and a random grid image selected from the eight sizes. Then, the model was trained to predict converted \gls{gt} which were formed by converting the real \gls{gt} into a grid-based \gls{gt} as in the previous section. 

The Unet model used dropout layers with the probability of $0.5$. Then, we used our Unet model as a stochastic model to perform Monte Carlo sampling for the validation data. We kept our Unet model in the training state to perform this sampling while predicting the output for the validation data. In the Pytorch library, which is used for all our implementations, we can do this simply by keeping the model state in the \texttt{model.train()} state. We iterated $50$ times for a single input to predict the output. We calculated the mean from these $50$ predictions, which is used as the final prediction for the competition and \gls{std} images to know the model's confidence for the predictions. The whole training process is illustrated in Figure \ref{fig:model} with an example image and a grid size of $8 \times 8$ as an input. However, we submitted the predicted mean images for the gird size of $ 256 \times 256$ which generate predictions with the size of true \gls{gt} (without any transformations).  All the experiments used a fixed learning rate of $0.001$ with the RMSprop optimizer~\cite{hinton2012neural}, which were selected from preliminary experiments. 




\section{Result and discussion}

\input{tables/results}

Table~\ref{tbl:results} summarizes the \gls{miou} and the \gls{dc} for the validation dataset and the test dataset. The final results to the competition were collected from mean images calculated by sampling $ 50 $ outputs for the same input with the grid size of $ 256 \time 256$. Additionally, we have calculated \gls{std} images for the validation dataset to show the benefits of using PYRA. Example outputs for a given input image are illustrated in Figure \ref{fig:main_grid}.

According to the results in Table~\ref{tbl:results}, Exp-3 which use only Pyramid-focus-augmentation shows the best validation results with \gls{miou} of $0.7693$ and \gls{dc} of $0.8447$, and the best test results with \gls{miou} of $0.6981$ and \gls{dc} of $0.7887$. The advantage of our Pyramid-focus-augmentation can be identified using the third row of Figure~\ref{fig:main_grid} along the fourth row of the same figure. We can see that our model can focus on polyp regions step by step. The third row of Figure~\ref{fig:main_grid} shows how our model predicts correct polyp cells in $ 2\times2, 4 \times 4, 8 \times 8, 16 \times 16, 32 \times 32, 64 \times 64 $, $ 128 \times 128 $  and $ 256\times256$ grid sizes, respectively. When we compare this row with the last row of the images of \gls{std}, we can see that the model has high confidence for the identified polyp regions. For example, it shows high confidence (black color region) for the middle part of the polyps. In contrast, our model shows less confidence (yellow color region) for a polyps' outer borders. 




\section{Conclusion and future work}

In this paper, we presented a novel augmentation method called Pyramid-focus-augmentation (PYRA), which can be used to train segmentation \gls{dl} methods. Our method shows a large benefit in the  medical diagnosis use-case, by focusing a doctors' attention on regions with findings step by step. 

Our experiments did not use post-processing to clean up output corresponding to the input grid. In future work, we will evaluate our approach with additional post-processing steps for smaller grid sizes. For example, we can do convolution operations to the output using a convolutional window equal to the input grid size to clean the results. However, post-processing techniques will not improve the final results when the grid size equals the input images' resolution. 
%

\section{Acknowledgment}
The research has benefited from the Experimental Infrastructure for Exploration of Exascale Computing (eX3), which is financially supported by the Research Council of Norway under contract 270053.

\newpage
\balance
\bibliographystyle{ACM-Reference-Format}
\bibliography{mybibfile} 

\end{document}

%% file: images/model.tex
\begin{figure}[!t]
    \centering
    \includegraphics[scale=0.6]{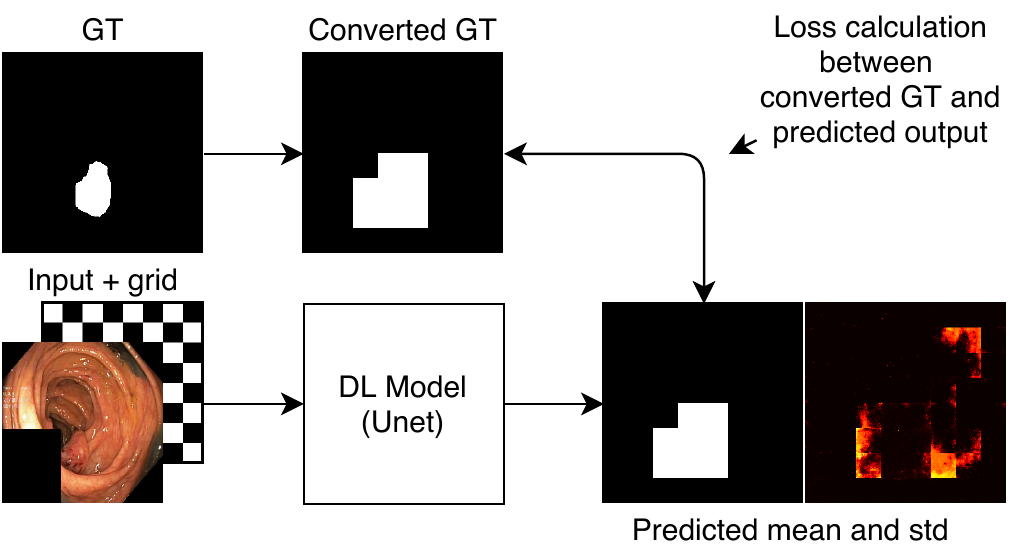}
    \caption{Training steps for a segmentation model with the new augmentation technique.}
    \label{fig:model}
    \vspace{-20pt}
\end{figure}

%% file: images/main_grid.tex
\begin{figure*}[!h]
  \centering
  \newcommand{\imagesize}{.06\linewidth}
  \newcommand{\imagefontsz}{\scriptsize}
  
    \begin{subfigure}[b]{\imagesize}
    \centering \imagefontsz Image
  \end{subfigure}
  \begin{subfigure}[b]{\imagesize}
  	\centering \imagefontsz $ 2\times 2 $
  \end{subfigure}
  \begin{subfigure}[b]{\imagesize}
    \centering \imagefontsz $ 4\times 4 $
  \end{subfigure}
  \begin{subfigure}[b]{\imagesize}
    \centering \imagefontsz $ 8\times 8 $
  \end{subfigure}
  \begin{subfigure}[b]{\imagesize}
    \centering \imagefontsz $ 16\times 16 $
  \end{subfigure}
  \begin{subfigure}[b]{\imagesize}
    \centering \imagefontsz $ 32\times 32 $
  \end{subfigure}
   \begin{subfigure}[b]{\imagesize}
    \centering \imagefontsz $ 64\times 64 $
  \end{subfigure}
  \begin{subfigure}[b]{\imagesize}
    \centering \imagefontsz $ 128\times 128 $
  \end{subfigure}
  \begin{subfigure}[b]{\imagesize}
    \centering \imagefontsz $ 256\times 256 $
  \end{subfigure}
  
  \vspace{3px}

  \begin{subfigure}[b]{\imagesize}
    \includegraphics[width=\linewidth]{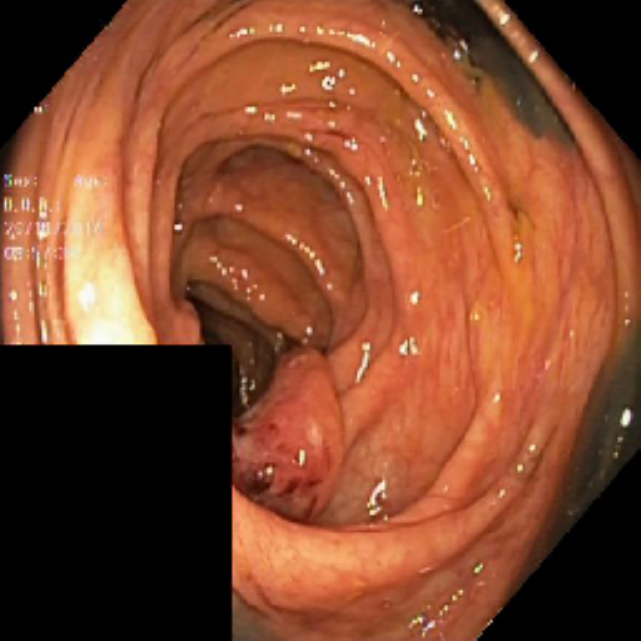}
  \end{subfigure}
  \begin{subfigure}[b]{\imagesize}
    \includegraphics[width=\linewidth]{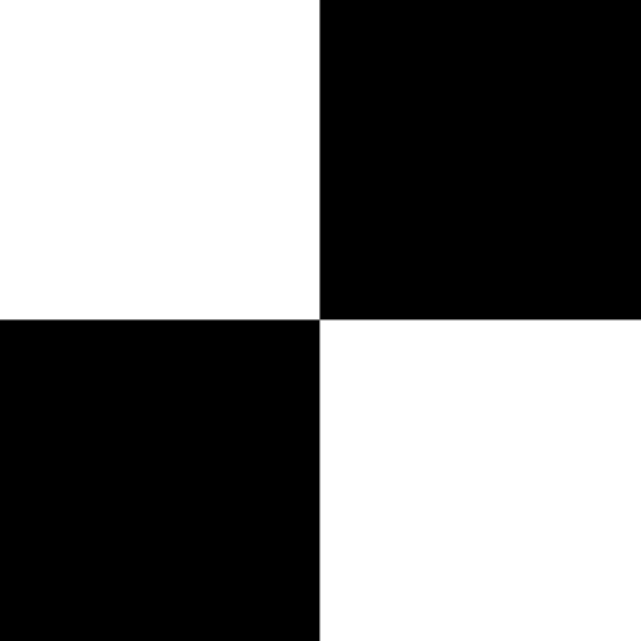}
  \end{subfigure}
  \begin{subfigure}[b]{\imagesize}
    \includegraphics[width=\linewidth]{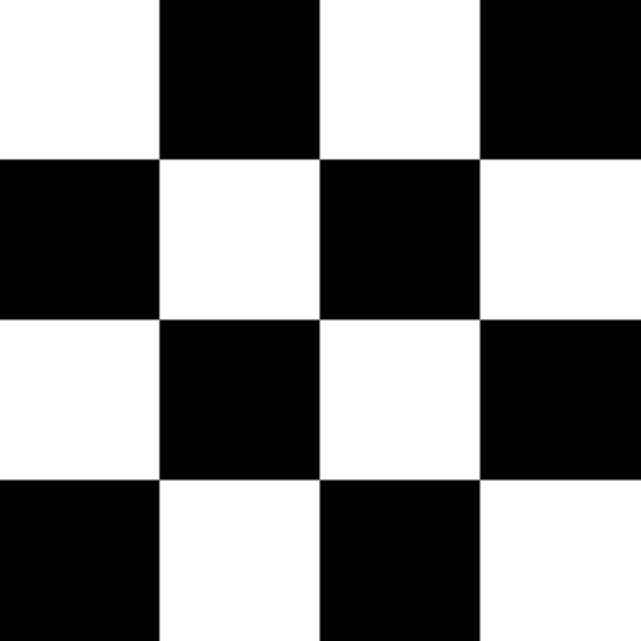}
  \end{subfigure}
  \begin{subfigure}[b]{\imagesize}
    \includegraphics[width=\linewidth]{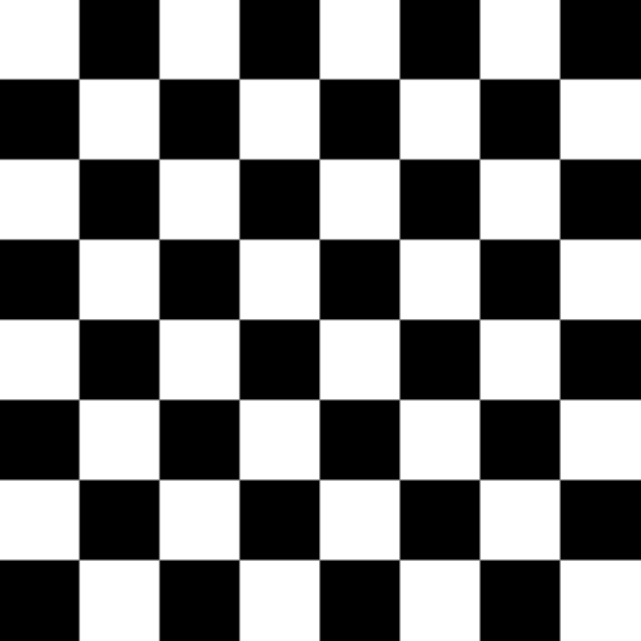}
  \end{subfigure}
  \begin{subfigure}[b]{\imagesize}
    \includegraphics[width=\linewidth]{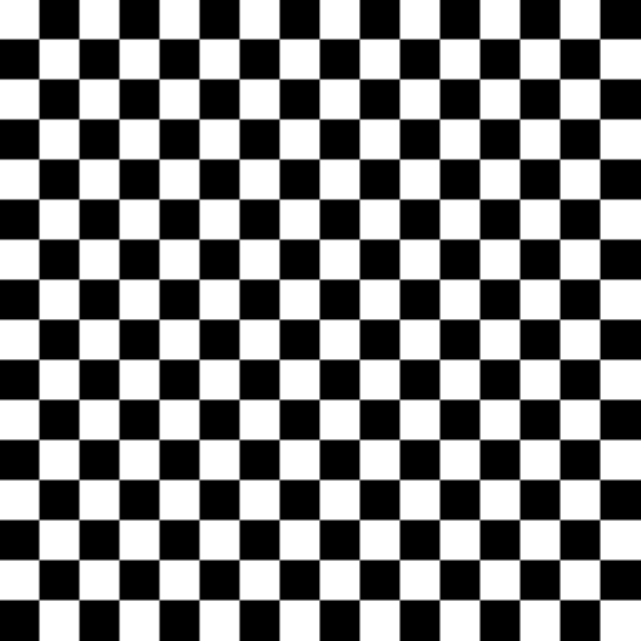}
  \end{subfigure}
  \begin{subfigure}[b]{\imagesize}
    \includegraphics[width=\linewidth]{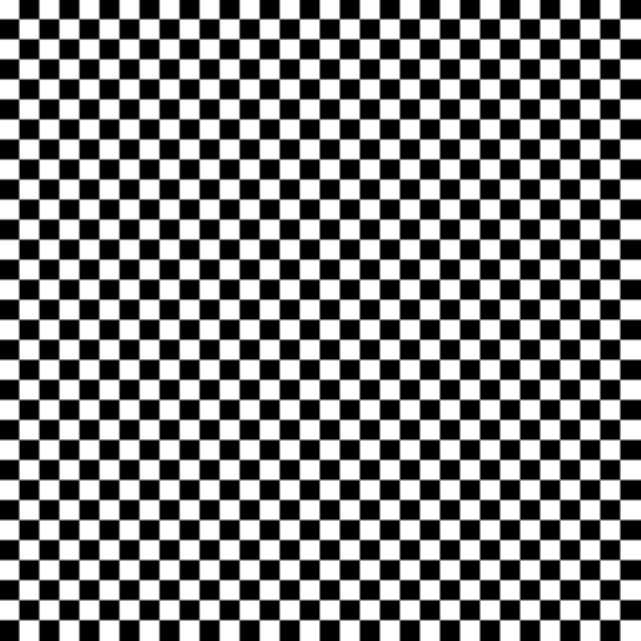}
  \end{subfigure}
  \begin{subfigure}[b]{\imagesize}
    \includegraphics[width=\linewidth]{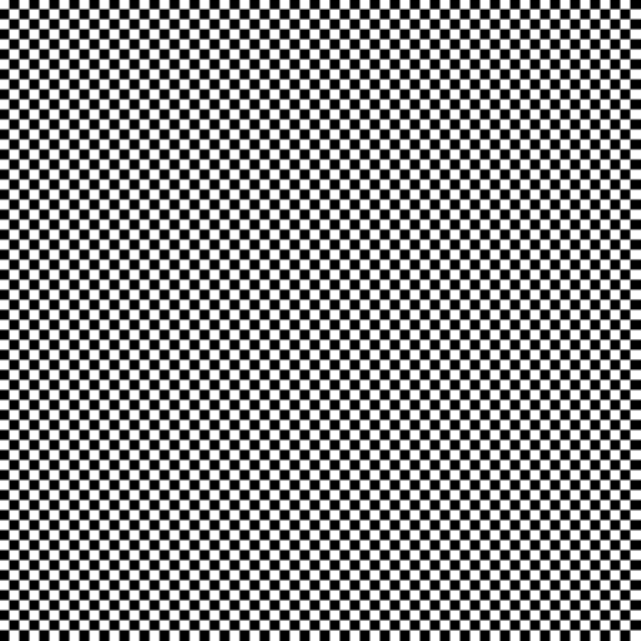}
  \end{subfigure}
  \begin{subfigure}[b]{\imagesize}
    \includegraphics[width=\linewidth]{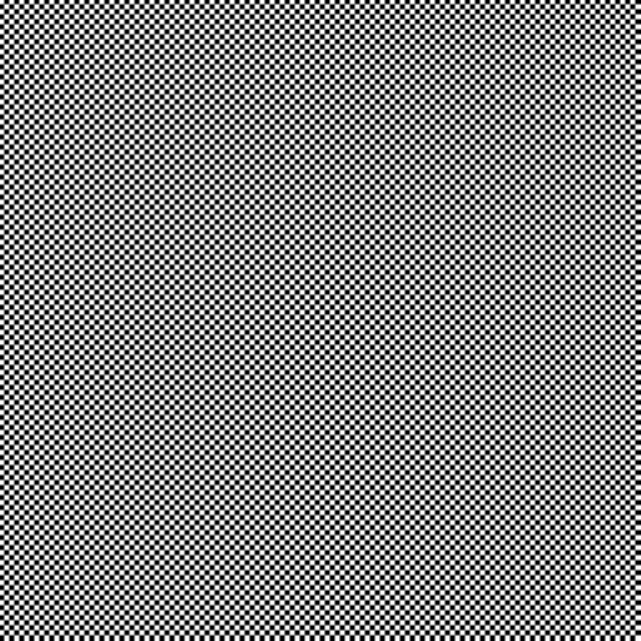}
  \end{subfigure}
  \begin{subfigure}[b]{\imagesize}
    \includegraphics[width=\linewidth]{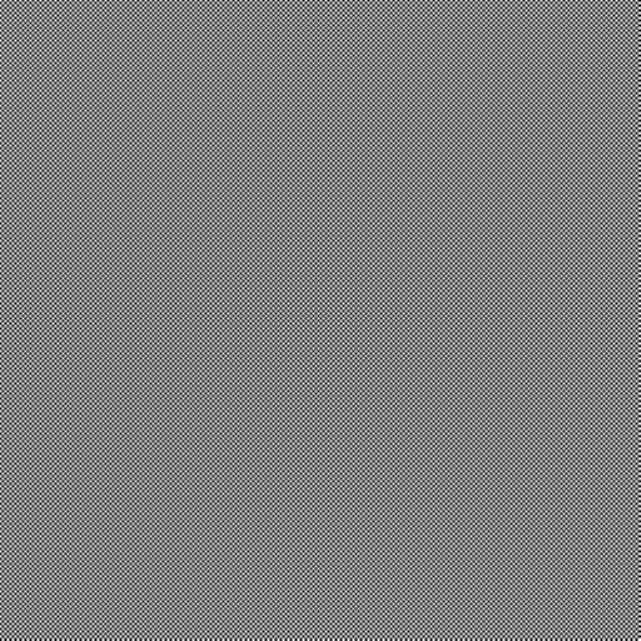}
  \end{subfigure}
  
  \vspace{3px}
  
    \begin{subfigure}[b]{\imagesize}
    \centering \imagefontsz Ground Truth
    \end{subfigure}
  \begin{subfigure}[b]{\imagesize}
    \includegraphics[width=\linewidth]{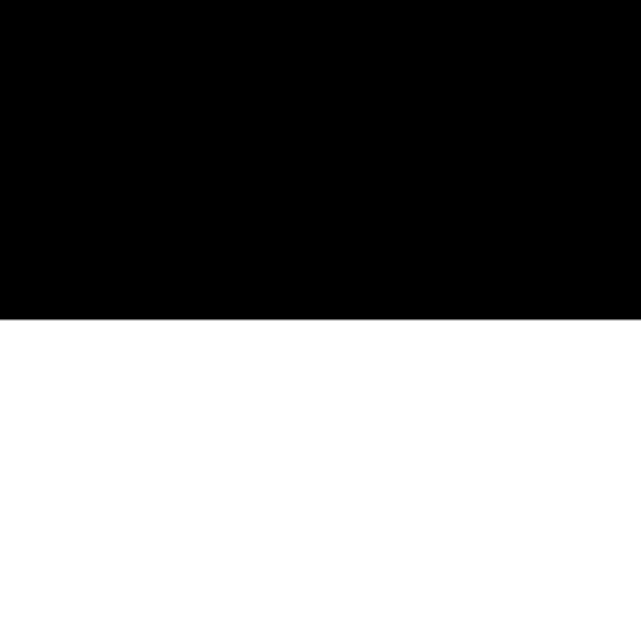}
  \end{subfigure}
  \begin{subfigure}[b]{\imagesize}
    \includegraphics[width=\linewidth]{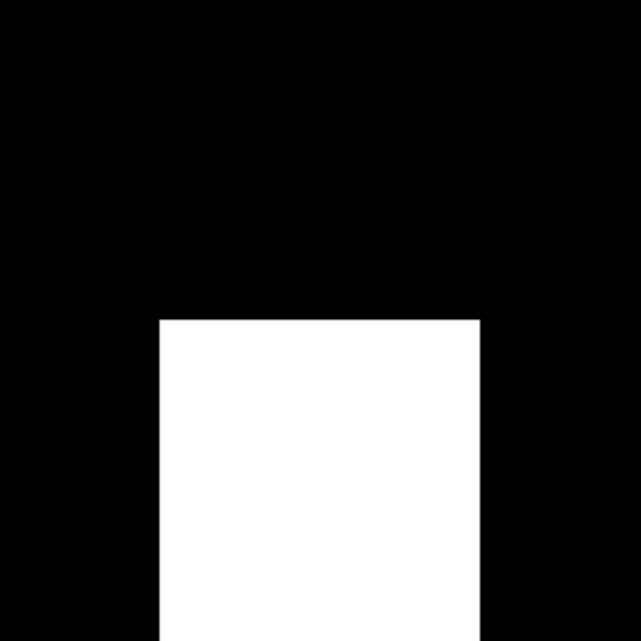}
  \end{subfigure}
  \begin{subfigure}[b]{\imagesize}
    \includegraphics[width=\linewidth]{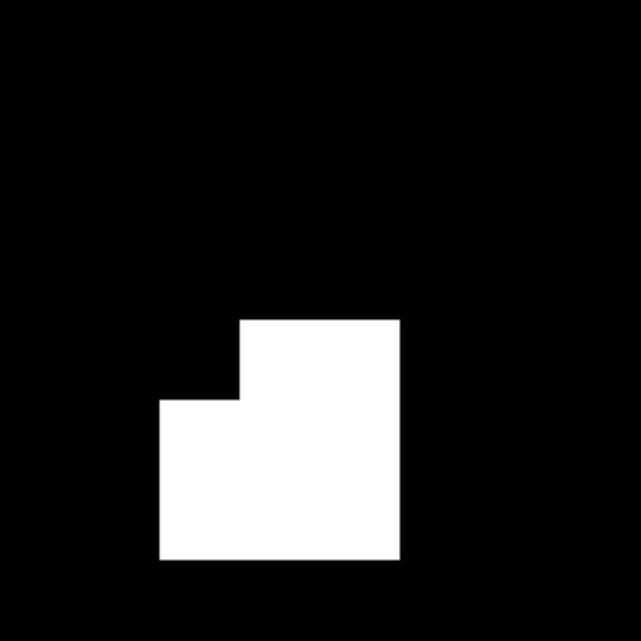}
  \end{subfigure}
  \begin{subfigure}[b]{\imagesize}
    \includegraphics[width=\linewidth]{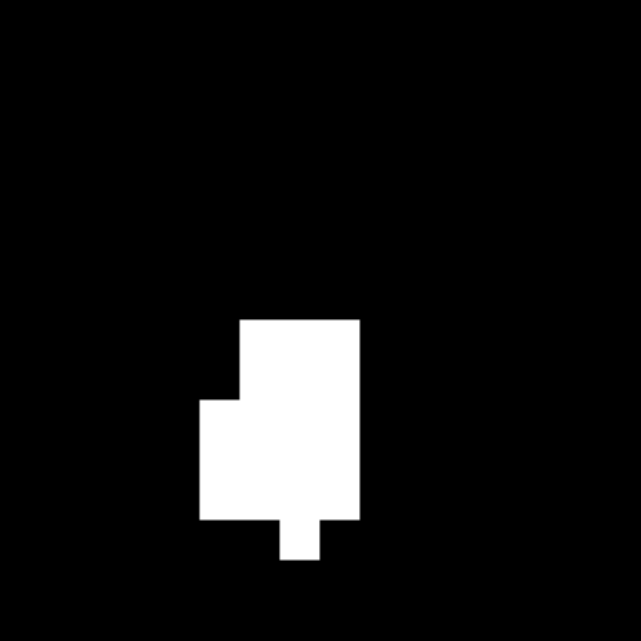}
  \end{subfigure}
  \begin{subfigure}[b]{\imagesize}
    \includegraphics[width=\linewidth]{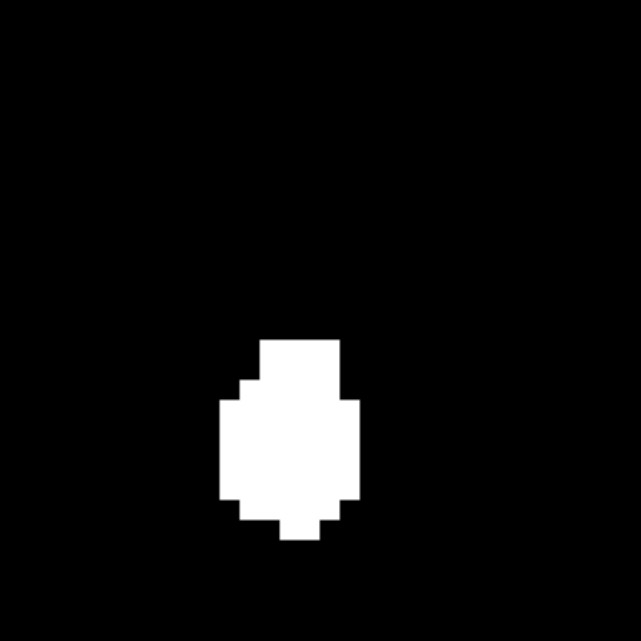}
  \end{subfigure}
  \begin{subfigure}[b]{\imagesize}
    \includegraphics[width=\linewidth]{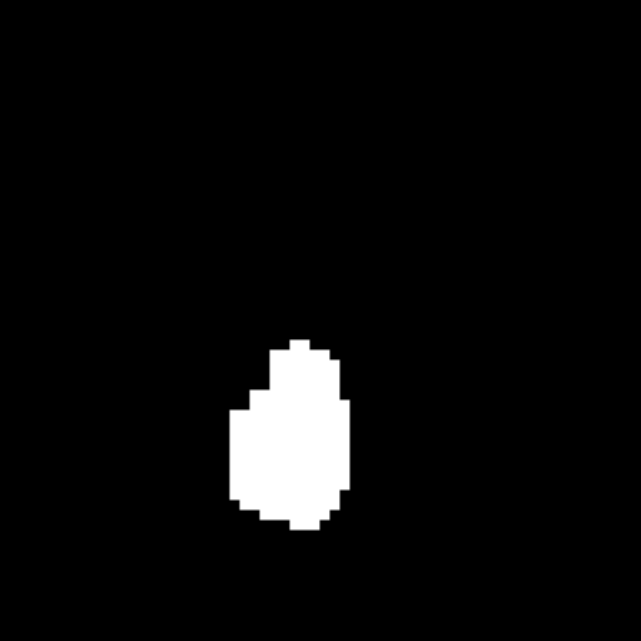}
  \end{subfigure}
  \begin{subfigure}[b]{\imagesize}
    \includegraphics[width=\linewidth]{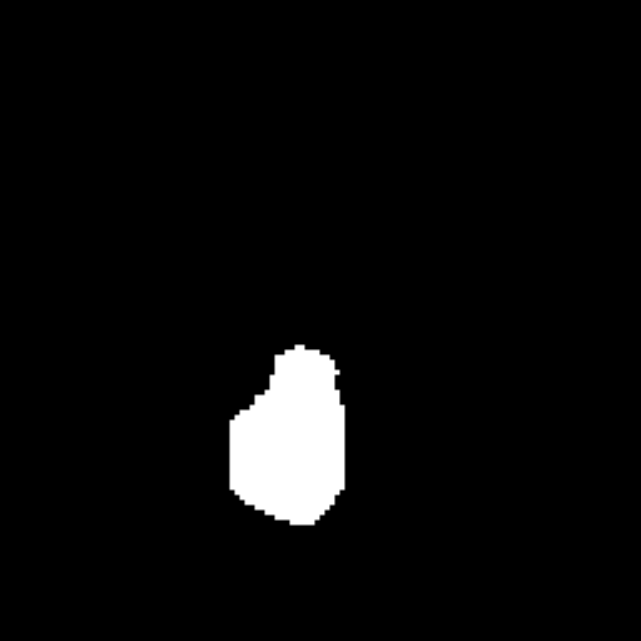}
  \end{subfigure}
  \begin{subfigure}[b]{\imagesize}
    \includegraphics[width=\linewidth]{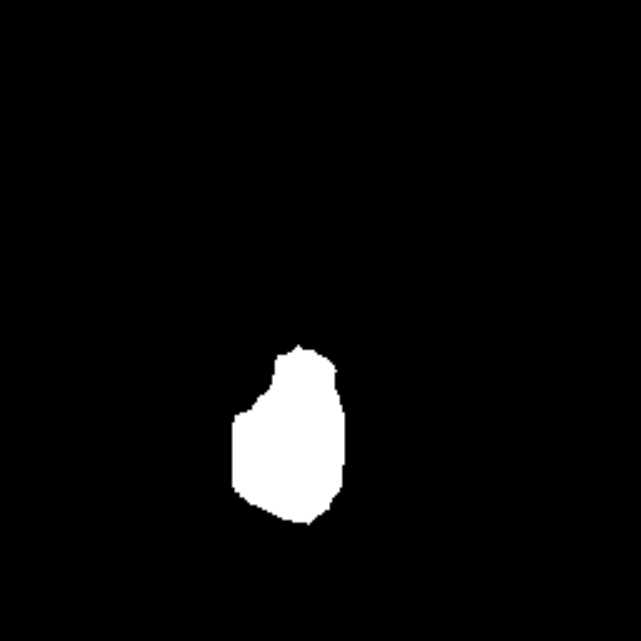}
  \end{subfigure}
  
  \vspace{3px}
  
     \begin{subfigure}[b]{\imagesize}
    \centering \imagefontsz Predictions
    \end{subfigure}
  \begin{subfigure}[b]{\imagesize}
    \includegraphics[width=\linewidth]{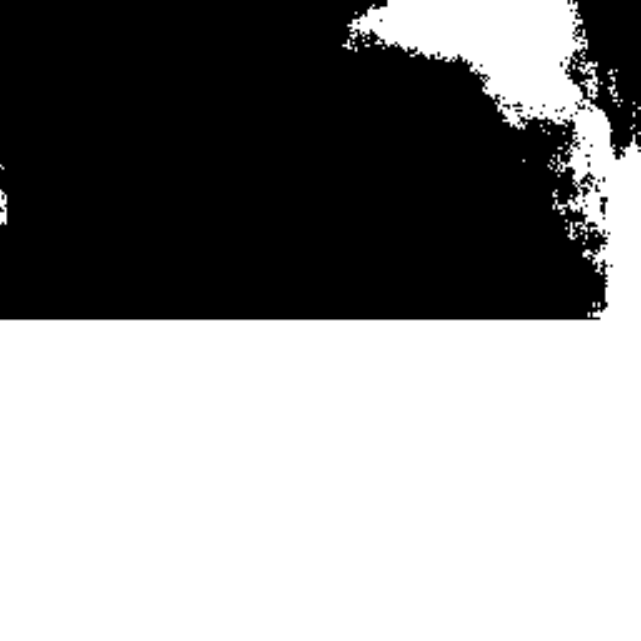}
  \end{subfigure}
  \begin{subfigure}[b]{\imagesize}
    \includegraphics[width=\linewidth]{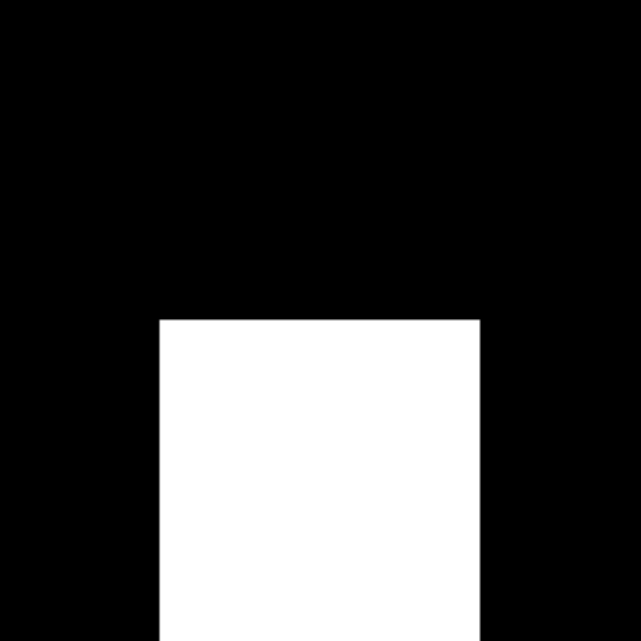}
  \end{subfigure}
  \begin{subfigure}[b]{\imagesize}
    \includegraphics[width=\linewidth]{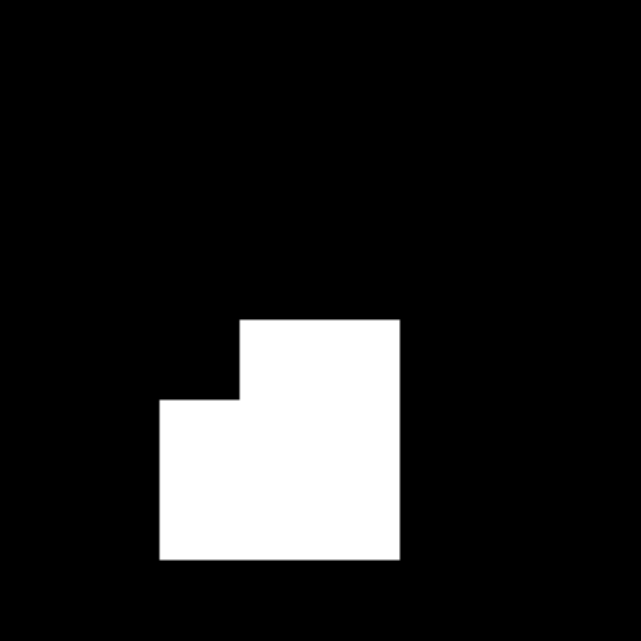}
  \end{subfigure}
  \begin{subfigure}[b]{\imagesize}
    \includegraphics[width=\linewidth]{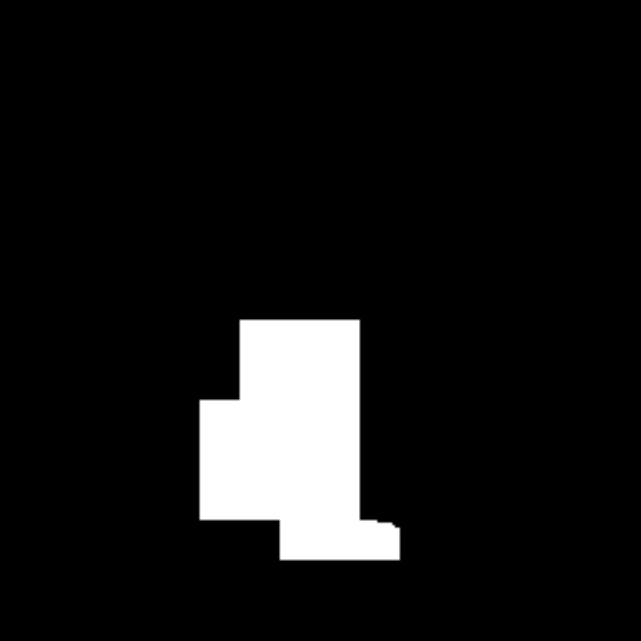}
  \end{subfigure}
  \begin{subfigure}[b]{\imagesize}
    \includegraphics[width=\linewidth]{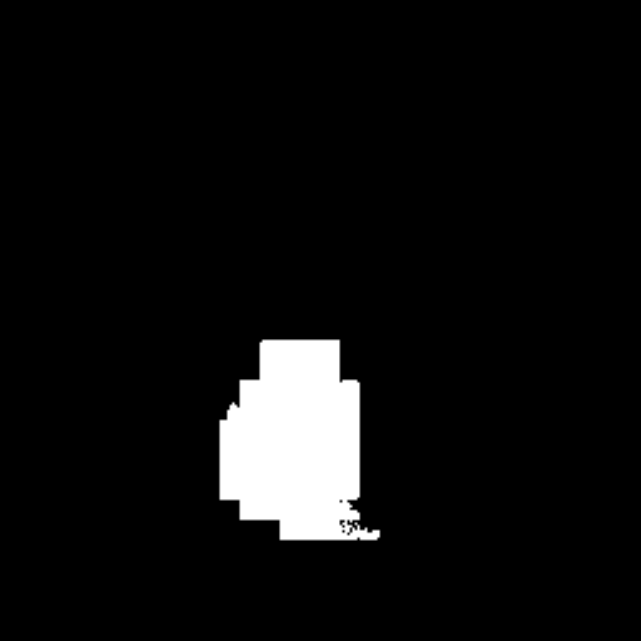}
  \end{subfigure}
  \begin{subfigure}[b]{\imagesize}
    \includegraphics[width=\linewidth]{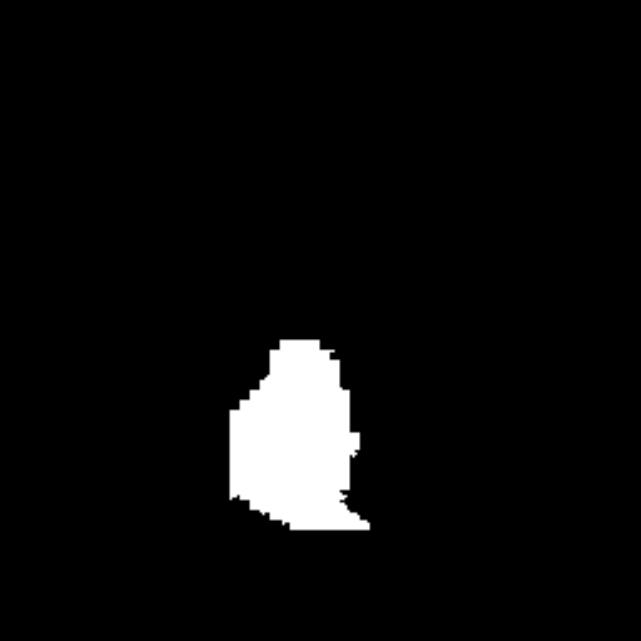}
  \end{subfigure}
  \begin{subfigure}[b]{\imagesize}
    \includegraphics[width=\linewidth]{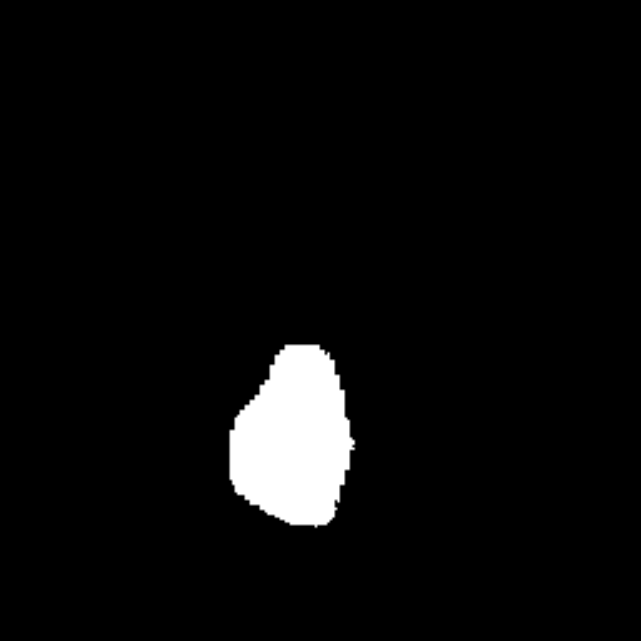}
  \end{subfigure}
  \begin{subfigure}[b]{\imagesize}
    \includegraphics[width=\linewidth]{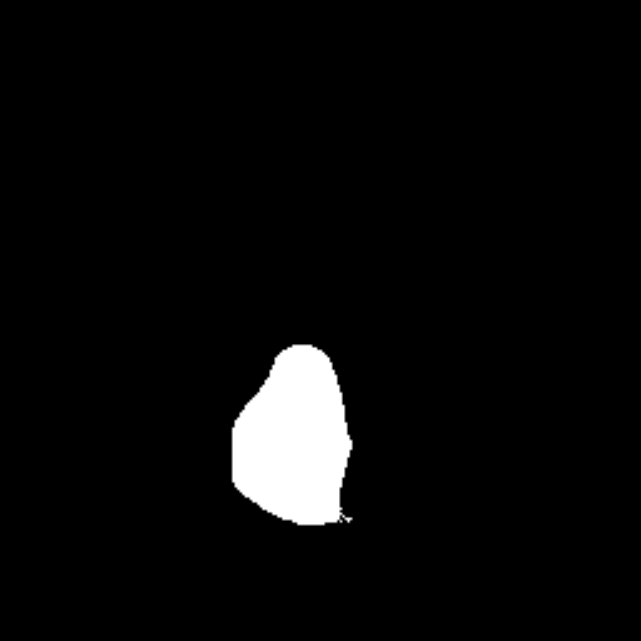}
  \end{subfigure}
  
  \vspace{3px}
  
     \begin{subfigure}[b]{\imagesize}
    \centering 
    \imagefontsz Std from 30 samples
    \end{subfigure}
  \begin{subfigure}[b]{\imagesize}
    \includegraphics[width=\linewidth]{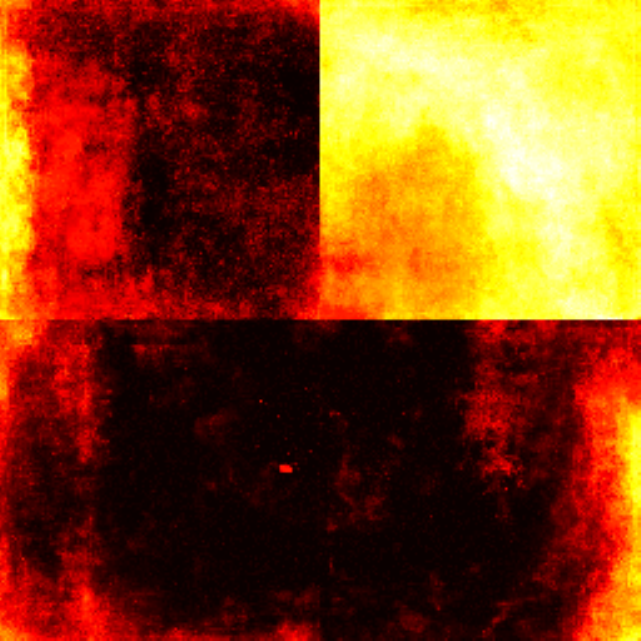}
  \end{subfigure}
  \begin{subfigure}[b]{\imagesize}
    \includegraphics[width=\linewidth]{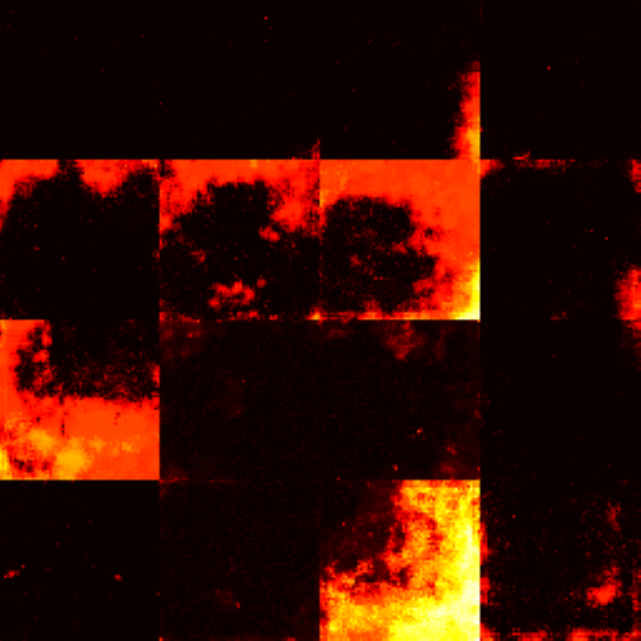}
  \end{subfigure}
  \begin{subfigure}[b]{\imagesize}
    \includegraphics[width=\linewidth]{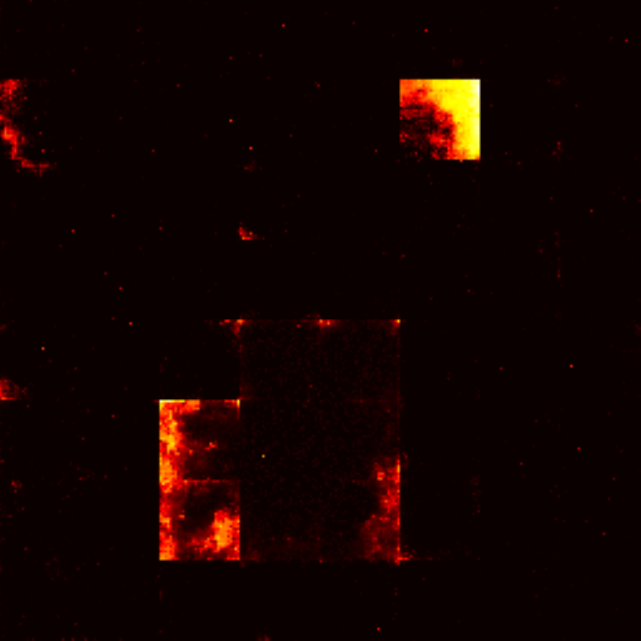}
  \end{subfigure}
  \begin{subfigure}[b]{\imagesize}
    \includegraphics[width=\linewidth]{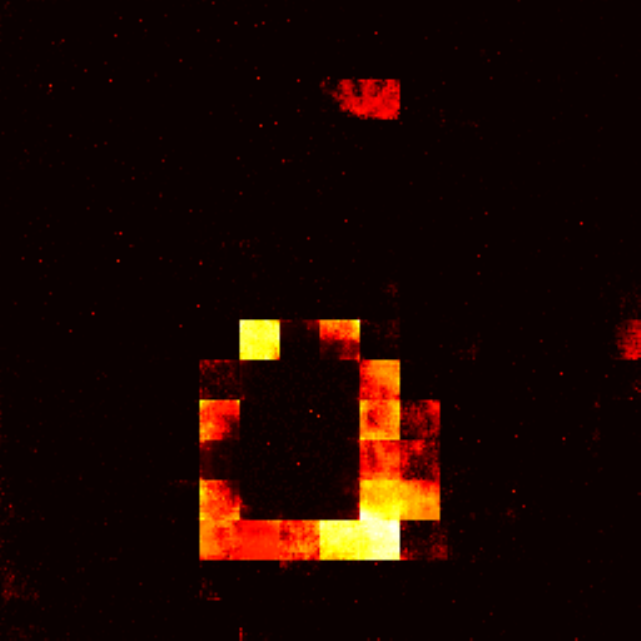}
  \end{subfigure}
  \begin{subfigure}[b]{\imagesize}
    \includegraphics[width=\linewidth]{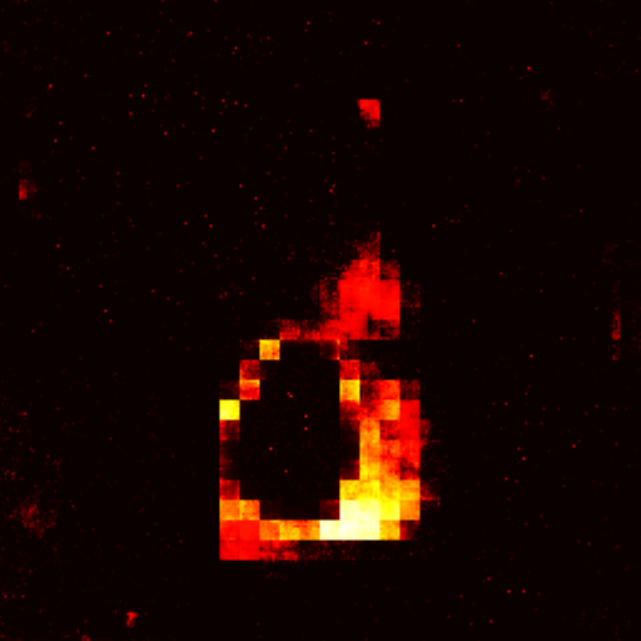}
  \end{subfigure}
  \begin{subfigure}[b]{\imagesize}
    \includegraphics[width=\linewidth]{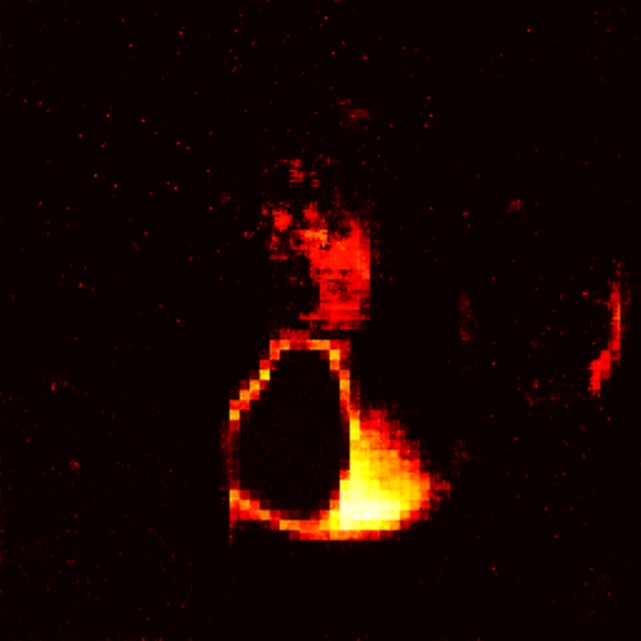}
  \end{subfigure}
  \begin{subfigure}[b]{\imagesize}
    \includegraphics[width=\linewidth]{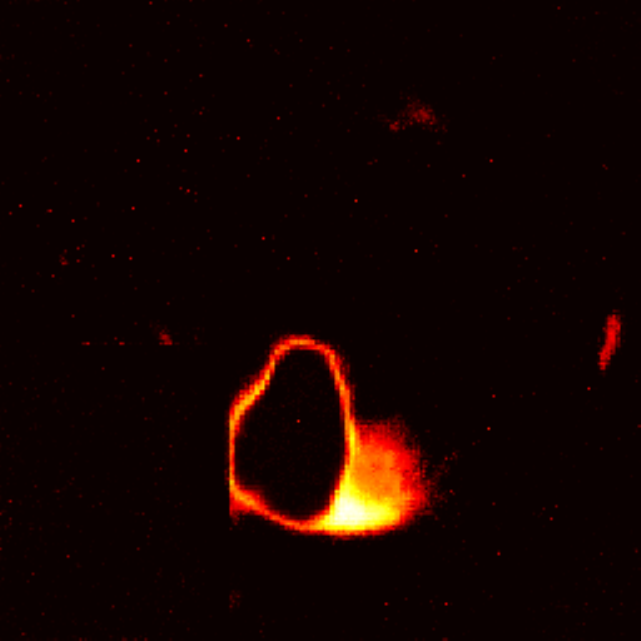}
  \end{subfigure}
  \begin{subfigure}[b]{\imagesize}
    \includegraphics[width=\linewidth]{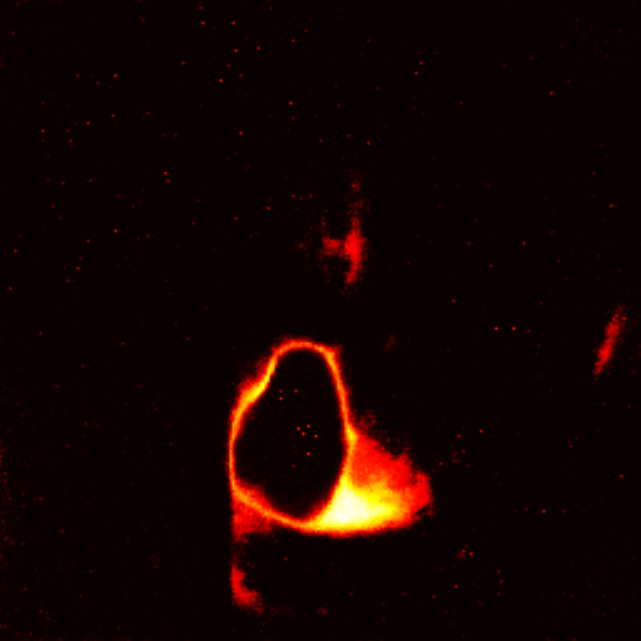}
  \end{subfigure}
  \vspace{-5pt}
  \caption{A representation of input and corresponding outputs of grid-augmentation-based segmentation. The first row shows an input image and all grid sizes used as stacked grid image with the input image. The second row represent ground truth. The third and fourth rows show predicted mean and std output images calculated from 30 samples.}
  \label{fig:main_grid}
  \vspace{-3mm}
\end{figure*}

%% file: tables/results.tex
\begin{table}[t!]
\caption{Result collected from validation data and test data. All test data results were provided by organizers of Medico task in MediaEval 2020.}
\begin{tabular}{ccccc}
\toprule
 & \multicolumn{2}{c}{Validation results} & \multicolumn{2}{c}{Official test results} \\
 \midrule
Method & mIOU & Dice & mIOU & Dice \\
\midrule
Exp-1 & 0.7640 & 0.8422 & 0.6934 & 0.7817 \\
Exp-2 & 0.7077 & 0.7957 & 0.6759 & 0.7700 \\
Exp-3 & \textbf{0.7693} & \textbf{0.8447} & \textbf{0.6981} & \textbf{0.7887}\\
Exp-4 & 0.6898 & 0.7822 & 0.6696 & 0.7665 \\
\bottomrule
\end{tabular}
\label{tbl:results}
\vspace{-20pt}
\end{table}